\documentclass[sigconf]{acmart}

\usepackage{booktabs} 
\usepackage{multirow}
\graphicspath{ {images/} }

\usepackage{url}
\usepackage{hyperref}
\usepackage{breakurl}

\usepackage{tikz}
\tikzset{%
    every neuron/.style={
        circle,
        draw,
        minimum size=1cm
    },
    neuron missing/.style={
        draw=none, 
        scale=4,
        text height=0.333cm,
        execute at begin node=\color{black}$\vdots$
    },
    every s_node/.style={ 
        circle,
        draw,
        minimum size=0.1cm
    },
    every bias/.style={ 
        rectangle,
        draw,
        minimum size=0.1cm
    },
}

\copyrightyear{2020}
\acmYear{2020}
\setcopyright{acmlicensed}\acmConference[GECCO '20]{Genetic and Evolutionary Computation Conference}{July 8--12, 2020}{Cancún, Mexico}
\acmBooktitle{Genetic and Evolutionary Computation Conference (GECCO '20), July 8--12, 2020, Cancún, Mexico}
\acmPrice{15.00}
\acmDOI{10.1145/3377930.3389843}
\acmISBN{978-1-4503-7128-5/20/07}

\begin{document}
\title{Towards Crossing the Reality Gap with Evolved Plastic Neurocontrollers}

\author{Huanneng Qiu}
\affiliation{%
  \institution{The University of New South Wales Canberra}
  \streetaddress{Northcott Drive, Campbell}
  \city{Canberra}
  \state{ACT}
  \country{Australia}
  \postcode{2600}
}
\email{huanneng.qiu@student.unsw.edu.au}

\author{Matthew Garratt}
\affiliation{%
  \institution{The University of New South Wales Canberra}
  \streetaddress{Northcott Drive, Campbell}
  \city{Canberra}
  \state{ACT}
  \country{Australia}
  \postcode{2600}
}
\email{m.garratt@adfa.edu.au}

\author{David Howard}
\orcid{1234-5678-9012}
\affiliation{%
  \institution{Robotics and Autonomous Systems Group, CSIRO}
  \streetaddress{1 Technology Court, Pullenvale}
  \city{Brisbane}
  \state{QLD}
  \country{Australia}
  \postcode{4069}
}
\email{david.howard@csiro.au}

\author{Sreenatha Anavatti}
\affiliation{%
  \institution{The University of New South Wales Canberra}
  \streetaddress{Northcott Drive, Campbell}
  \city{Canberra}
  \state{ACT}
  \country{Australia}
  \postcode{2600}
}
\email{s.anavatti@adfa.edu.au}

\renewcommand{\shortauthors}{H. Qiu et al.}

\begin{abstract}

A critical issue in evolutionary robotics is the transfer of controllers learned in simulation to reality. 
This is especially the case for small Unmanned Aerial Vehicles (UAVs), as the platforms are highly dynamic and susceptible to breakage.
Previous approaches often require simulation models with a high level of accuracy, otherwise significant errors may arise when the well-designed controller is being deployed onto the targeted platform.
Here we try to overcome the transfer problem from a different perspective, by designing a spiking neurocontroller which uses synaptic plasticity to cross the reality gap via online adaptation.
Through a set of experiments we show that the evolved plastic spiking controller can maintain its functionality by self-adapting to model changes that take place after evolutionary training,
and consequently exhibit better performance than its non-plastic counterpart.

\end{abstract}

%
%



\maketitle

\section{Introduction} \label{intro}

Unmanned Aerial Vehicles (UAVs) are challenging platforms for developing and testing advanced control techniques, because they are highly dynamic, with strong couplings between different subsystems \cite{Ng2004}.
Controller design for these agile platforms is naturally difficult, as a poorly-performing controller can lead to catastrophic consequences, e.g., the UAV crashing.
In addition, many learning approaches require large numbers of fitness evaluations.
Therefore, there still exist a large group of aerial robotic studies relying on simulations as an intermediate step to develop control algorithms \cite{Kendoul2012}. 

When simulating, it is not uncommon to derive UAV models mathematically from first principles \cite{Pounds2010,Alaimo2013}.
However, such models are ill-suited to capturing every aspect of the system dynamics, 
because some of them cannot easily be modeled analytically, e.g., actuator kinematic nonlinearities, servo dynamics, etc \cite{Garratt2012}.
Ignoring these effects can significantly deteriorate the performance of the designed controller when being deployed onto the targeted platform.
To address this issue, a common practice is to develop control algorithms based on an `identified' model that is a simulated representation of the real plant.
This identified model is obtained by applying a data-driven process called \emph{system identification} that models the exact dynamics from the measured plant's input and output data. 
Such implementations have been successful amongst previous research \cite{Ng2004, Ng2006, Garratt2012, Kendoul2012, Hoffer2014}. 

While a lot of works have pursued a perfect model that well characterizes UAV platforms,
a key issue is that loss of performance is still likely to happen when transferring the well-designed (in simulation) controller onto the real platform that has somewhat different dynamics -- the well-known {\em reality gap}.
In this work we demonstrate a novel approach to compensate the gap across different platform representations, which works specifically with Spiking Neural Networks (SNNs) that exhibit online adaptation ability through Hebbian plasticity \cite{Gerstner2002}.
We propose an evolutionary learning strategy for SNNs, which includes topology and weight evolution as per NEAT \cite{Stanley2002}, and integration of biological plastic learning mechanisms.
With the goal of simulation-to-reality transfer, we here prove the concept in a time-efficient manner by transferring from a simpler to a more complex model, a transfer that encapsulates some issues inherent in crossing the reality gap, i.e. incomplete capture of true flight dynamics, oversimplification of true conditions.

In this work, we focus on the development of UAV height control.
Our approach to resolve this problem is threefold.
First, explicit mathematical modeling of the aircraft is not required. Instead, a simplified linear model is identified based on the measurement of the plant's input and output data.
In reality, such models are fast to run and simple to develop.
Second, neuroevolution takes place as usual to search through solution space for the construction of high-performance networks.
Finally, Hebbian plasticity is implemented by leveraging evolutionary algorithms to optimize plastic rule coefficients that describe how neural connections are updated. 
Plasticity evolution has been used in conventional ANNs \cite{Urzelai2001, Soltoggio2007, Tonelli2011} and SNNs \cite{Howard2012a},
where evolution takes place in the rules that govern synaptic self-organization instead of in the synapses themselves.
The evolved controller is able to exhibit online adaptation due to plasticity, which allows successful transfer to a more realistic model and indicates that transfer to reality would be similarly successful.

Organization of the rest of this paper is as follows. 
Section \ref{sect:snn} introduces our SNN package that is utilized to develop our UAV controller, including descriptions of spiking neuron models, the mechanism of plasticity learning and evolutionary learning strategies.
Section \ref{sect:sysmdl} presents the plant model to be controlled in this work.
Section \ref{sect:prob_desc}, \ref{sect:sysid} and \ref{sect:ctrller} describe the controller development process in detail.
Results and analysis are given in Section \ref{sect:results}.
Finally, discussions and conclusions are presented in Section \ref{sect:discuss} and \ref{sect:conclusion}.


\section{eSpinn: Learning with Spikes} \label{sect:snn}

\subsection{Background}

The current widely used Artificial Neural Networks (ANNs) follow a computation cycle of \emph{multiply-accumulate-activate}.
The neuron model consists of two components: a weighted sum of inputs and an activation function generating the output accordingly. 
Both the inputs and outputs of these neurons are real-valued. 
While ANN models have shown exceptional performance in the artificial intelligence domain, they are highly abstracted from their biological counterparts in terms of information representation, transmission and computation paradigms.

SNNs, on the other hand, carry out computation based on biological modeling of neurons and synaptic interactions,
and have been of great interest in the computational intelligence community in recent decades.
Applications have been both non-behavioral \cite{Abbott2016} and behavioral \cite{Vasu2017, Qiu2018}.
Information transmission in SNNs is by means of discrete \emph{spikes} generated during a potential integration process.
Such spatiotemporal dynamics are able to yield more powerful computation compared with non-spiking neural systems \cite{Maass1997}.
Moreover, neuromorphic hardware implementations of SNNs are believed to provide fast and low-power information processing due to their event-driven sparsity \cite{Bouvier2019},
which perfectly suits embedded applications such as UAVs.

\begin{figure}[htbp]
    \centering
    \includegraphics[width=0.33\textwidth]{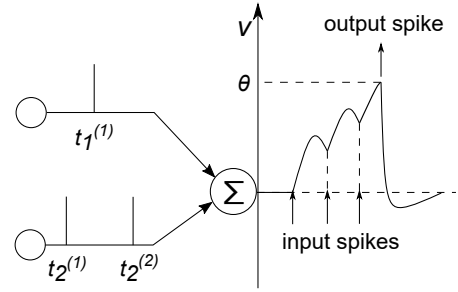}
    \caption{Illustration of spike transmission in SNNs.
    Membrane potential $v$ accumulates as input spikes arrive and decays with time.
    Whenever it reaches a given threshold $\theta$, an output spike will be fired,
    and the potential will be reset to a resting value.}
    \label{fig:illus}
\end{figure}

As shown in Fig. \ref{fig:illus}, spikes are fired at certain points in time, whenever the \emph{membrane potential} of a neuron exceeds its threshold. 
They will travel through synapses from the \emph{presynaptic} neurons and arrive at all forward-connected \emph{postsynaptic} neurons. 
The information measured by spikes is in form of timing and frequency, rather than the amplitude or intensity.

In order to assist the process of designing our spiking controller, we have developed the \texttt{eSpinn} software package. 
The \texttt{eSpinn} library stands for \textbf{E}volving \textbf{Spi}king \textbf{N}eural \textbf{N}etworks. 
It is designed to develop controller learning strategies for nonlinear control models by integrating biological learning mechanisms with neuroevolution algorithms. 
It is able to accommodate different network implementations (ANNs, SNNs and hybrid models) with specific dataflow schemes.
\texttt{eSpinn} is written in C++ and has abundant interfaces to easily archive data through serialization.
It also contains scripts for data visualization and integration with MATLAB and Simulink simulations.

\subsection{Neuron Model}

To date, there have been different kinds of spiking neuron models. 
When implementing a neuron model, trade-offs must be considered between biological reality and computational efficiency.
In this work we use the two-dimensional Izhikevich model \cite{Izhikevich2003}, 
because of its capability of exhibiting richness and complexity in neuron firing behavior (detailed in \cite{Izhikevich2003}) with only two ordinary differential equations:

\begin{equation}
\begin{aligned}
\dot{v} &= 0.04v^2 + 5v +140 -u + I \\
\dot{u} &= a(bv - u)
\end{aligned}
\label{eq:izhi}
\end{equation}
with after-spike resetting following:
\begin{equation}
\text{if } v \geq v_t \text{, then}
\left \{
    \begin{array}{l}
        v =c \\
        u = u + d
    \end{array}
\right.
\label{eq:spikereset}
\end{equation}

\begin{figure}[htbp]
    \centering
    \includegraphics[width=0.47\textwidth]{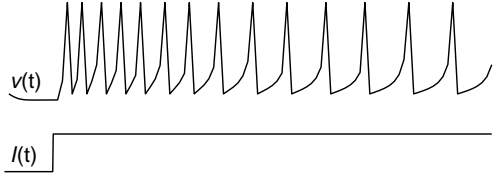}
    \caption{Membrane potential response $v(t)$ to an external current signal $I(t)$ of an Izhikevich neuron with the following settings: $a$ = 0.02; $b$ = 0.2; $c$ = -65; $d$ = 2.}
    \label{fig:izhi}
\end{figure}

Here $v$ represents the membrane potential of the neuron; $u$ represents a recovery variable; $\dot{v}$ and $\dot{u}$ denote their derivatives, respectively.
$I$ represents the synaptic current that is injected into the neuron.
Whenever $v$ exceeds the threshold of membrane potential $v_t$, a spike will be fired and $v$ and $u$ will be reset following Eq.~\eqref{eq:spikereset}.
$a, b, c$ and $d$ are dimensionless coefficients that are tunable to form different firing patterns \cite{Izhikevich2003}. 
The membrane potential response of an Izhikevich neuron is given in Fig. \ref{fig:izhi}, with an injected current signal.

A spike train is defined as a temporal sequence of firing times: 
\begin{equation}
s(t) = \sum_f \delta(t-t^{(f)})
\end{equation}
where $\delta(t)$ is the Dirac $\delta$ function;
$t^{(f)}$ represents the firing time, i.e., the moment of $v$ crossing threshold $v_t$ from below.

\subsection{Network Structure}

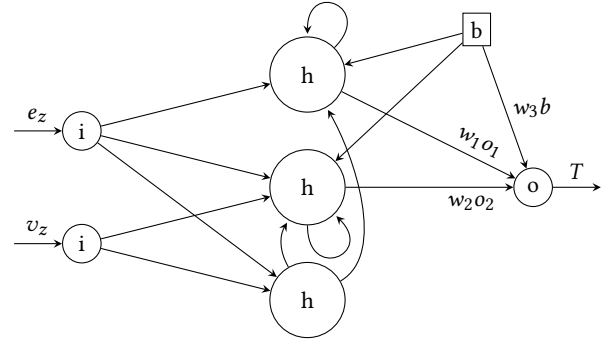
\begin{figure}[htbp]
    \centering
    \begin{tikzpicture}[x=1.5cm, y=1.5cm, >=stealth]

    \foreach \m/\l [count=\y] in {1,2}
    \node [every s_node/.try, innode \m/.try] (input-\m) at (0,0.5+\y) {i};

    \foreach \m [count=\y] in {1,2,3}
    \node [every neuron/.try, neuron \m/.try ] (hidden-\m) at (2,\y) {h};

    \foreach \m [count=\y] in {1}
    \node [every s_node/.try, outnode \m/.try] (output-\m) at (4,1+\y) {o};

    \foreach \m [count=\y] in {1}
    \node [every bias/.try, bias \m/.try ] (bias-\m) at (3.5,2.4+\y) {b};

    \draw [<-] (input-1) -- ++(-0.6,0)
    node [above, midway] {$v_z$};

    \draw [<-] (input-2) -- ++(-0.6,0)
    node [above, midway] {$e_z$};


    \draw [->] (output-1) -- ++(0.6,0)
    node [above, midway] {$T$};

    \foreach \i in {1,...,2}
    \foreach \j in {1,...,2}
    \draw [->] (input-\i) -- (hidden-\j);
    \draw [->] (input-2) -- (hidden-3);

    \draw [->] (hidden-2) -- (output-1) 
    node [below, near end, sloped] {$w_2 o_2$};
    \draw [->] (hidden-3) -- (output-1) 
    node [above, near end, sloped] {$w_1 o_1$};

    \draw [->] (bias-1) -- (output-1)
    node [right, midway] {$w_3 b$};
    \draw [->] (bias-1) -- (hidden-2);
    \draw [->] (bias-1) -- (hidden-3);

    \draw[->,shorten >=1pt] (hidden-2) to [out=270,in=315,loop,looseness=5] (hidden-2);
    \draw[->,shorten >=1pt] (hidden-3) to [out=45,in=90,loop,looseness=5] (hidden-3);

    \draw[->,shorten >=1pt] (hidden-1) to [out=120,in=-120,loop,looseness=1] (hidden-2);
    \draw[->,shorten >=1pt] (hidden-1) to [out=30,in=-60,loop,looseness=0.8] (hidden-3);

    \end{tikzpicture}
    \caption{Spiking network topology that allows internal recurrence. 
    Network inputs ($i$) consist of position error in z-axis $e_z$ and vertical velocity $v_z$.
    Hidden layer neurons ($h$) are spiking, whose outputs $o_i$ involve direct transfer of intermediate membrane potential and decoding of firing rate.
    A bias neuron ($b$) is allowed to connect to any neurons in the hidden and output layer.
    Output thrust command $T$ is calculated based on a weighted sum of incoming neuron activations $\sum w_i o_i$, which will be fed to the hexacopter plant model.
    Weights $w_i$ are bounded within [-1, 1].}
    \label{fig:snn_topo}
\end{figure}

We use a three-layer architecture that has hidden-layer recurrent connections, illustrated in Fig. \ref{fig:snn_topo}.
The input layer consists of \emph{encoding} neurons which act as information converters.
Hidden-layer spiking neurons are connected via unidirectional weighted synapses among themselves. 
Such internal recurrence ensures a history of recent inputs is preserved within the network, which exhibits highly nonlinear functionality.
Output neurons can be configured as either activation-based or spiking. 
In this work a linear unit is used to obtain real-value outputs from a weighted sum of outputs from hidden-layer neurons.
A bias neuron that has a constant output value is able to connect to any neurons in the hidden and output layers. 
Connection weights are bounded within [-1, 1].
The NEAT topology and weight evolution scheme is used to form and update network connections and consequently to seek functional network compositions.

In a rate coding scheme, neuron output is defined as the spike train frequency calculated within a given time window.
Loss of precision during this process is likely to happen. 
\texttt{eSpinn} configures a decoding method with high accuracy to derive continuous outputs from discrete spike trains.
The implementation involves direct transfer of intermediate membrane potentials as well as decoding of spikes in a rate-based manner.

\subsection{Hebbian Plasticity}
\label{ssect:hebb}

In neuroscience, studies have shown that synaptic strength in biological neural systems is not fixed but changes over time \cite{Kandel1992} -- connections between pre- and postsynaptic neurons are updated according to their degree of causality,
which involves changes of synaptic weights or even formation/removal of synapses.
This phenomenon is often referred to as Hebbian plasticity as inspired by the Hebb's postulate \cite{Hebb1949}.
In our work, plastic behaviors are determined by leveraging evolutionary algorithms to optimize plastic rule coefficients, such that each connection is able to develop its own plastic rule.

Modern Hebbian rules generally describe weight change $\Delta w$ as a function of the joint activity of pre- and postsynaptic neurons:
\begin{equation}
    \Delta w = f(w_{ij}, u_j, u_i)
\end{equation}
where $w_{ij}$ represents the weight of the connection from neuron $j$ to neuron $i$; $u_j$ and $u_i$ represent the firing activity of $j$ and $i$, respectively.

In a spike-based scheme, we consider the synaptic plasticity at the level of individual spikes. 
This has led to a phenomenological temporal Hebbian paradigm: Spiking-Timing Dependent Plasticity (STDP) \cite{Gerstner2002}, which modulates synaptic weights between neurons based on the temporal difference of spikes.

While different STDP variants have been proposed \cite{Izhikevich2003a}, the basic principle of STDP is that the change of weight is driven by the causal correlations between the pre- and postsynaptic spikes. 
Weight change would be more significant when the two spikes fire closer together in time. 
The standard STDP learning window is formulated as:

\begin{equation}
W (\Delta t) =
\left \{
    \begin{array}{ll}
        A_+ e^{-\frac{\Delta t}{\tau_+}} & \Delta t > 0, \\
        A_- e^{\frac{\Delta t}{\tau_-}} & \Delta t < 0.
    \end{array}
\right.
\label{stdp}
\end{equation}
where $A_+$ and $A_-$ are scaling constants of strength of potentiation and depression; $\tau_+$ and $\tau_-$ represent the time decay constants; $\Delta t$ is the time difference between pre- and post-synaptic firing timings:
\begin{equation}
    \Delta t = t_{post} - t_{pre}
\end{equation}

\begin{figure}[htbp]
    \centering
    \includegraphics[width=0.44\textwidth]{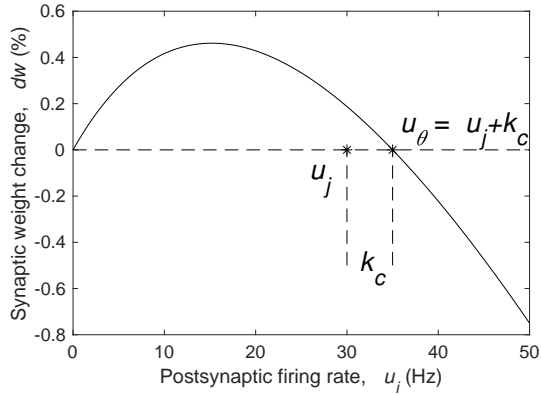}
    \caption{Hebbian learning curve with $A_+$ = 0.1, $A_-$ = -0.1, $\tau_+$ = 0.02\,s, $\tau_-$ = 0.02\,s}
    \label{fig:hebb}
\end{figure}

In \texttt{eSpinn} we have introduced a rate-based Hebbian model derived from the nearest neighbor STDP implementation \cite{Izhikevich2003a}, with two additional evolvable parameters:
\begin{equation}
    \dot{w} = u_i (\frac{A_+}{\tau_+^{-1}+u_i} + \frac{k_m (u_j-u_i+k_c)+A_-}{\tau_-^{-1}+u_i})
    \label{eq:hebb}
\end{equation}
where $k_m$ is a magnitude term that determines the amplitude of weight changes, and $k_c$ is a correlation term that determines the correlation between pre- and postsynaptic firing activity.
These factors are set to as evolvable so that the best values can be autonomously located.
Fig. \ref{fig:hebb} shows the resulting Hebbian learning curve. 
The connection weight has a stable converging equilibrium at $u_{\theta}$, which is due to the correlation term $k_c$.
This equilibrium corresponds to a balance of the pre- and postsynaptic firing.

\subsection{Learning in Neuroevolution} \label{neat}

While gradient methods have been very successful in training traditional MLPs \cite{Demuth2014}, 
their implementations on SNNs are not as straightforward because they require the availability of gradient information.
Instead, \texttt{eSpinn} has developed its own version of a popular neuroevolution approach -- NEAT \cite{Stanley2002}, which can accommodate different network implementations and integrate with Hebbian plasticity, as the method to learn the best network controller.

NEAT is a popular neuroevolution algorithm that involves network topology and weight evolution. 
It enables an incremental network topological growth to discover the (near) minimal effective network structure.

The basis of NEAT is the use of \emph{historical markings}, which are essentially gene IDs. 
They are used as a measurement of the genetic similarity of network topology, based on which, genomes are clustered into different species. 
Then NEAT uses an explicit fitness sharing scheme \cite{Eiben2015} to preserve network diversities.
Meanwhile, these markings are also used to line up genes from variant topologies and allow crossover of divergent genomes in a rationale manner.

\texttt{eSpinn} keeps a global list of innovations (e.g., structural variations), so that when an innovation occurs, we can know whether it has already existed. 
This mechanism will ensure networks with the same topology will have the exactly same innovation numbers, which is essential during the process of network structural growth.

\section{System Modeling} \label{sect:sysmdl}

The experimental platform is a commercial hexacopter, Tarot 680 Pro, fitted with a Pixhawk 2 autopilot system.
To assist the development and tests of our control paradigms, we have developed a Simulink model based on our previous work \cite{Santoso2017}.
The model is derived from first principles, which contains 6-DOF rigid body dynamics and non-linear aerodynamics.
Many aspects of the hexacopter dynamics are modeled with C/C++ S-functions, which describe the functionalities of Simulink blocks in C/C++ with MATLAB built-in APIs.

\begin{figure*}[htbp]
    \centering
    \includegraphics[width=0.83\textwidth]{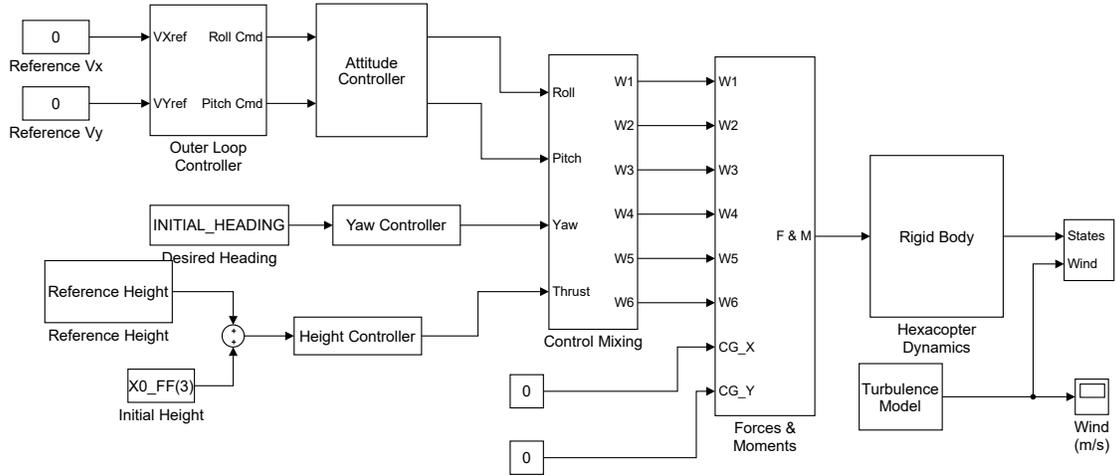}
    \caption{Top-level diagram of the hexacopter control model}
    \label{fig:hexa-diagram}
\end{figure*}

The simulation system is based on a hierarchical architecture. 
The top-level diagram of the system is given in Fig.~\ref{fig:hexa-diagram}. 
The `Control Mixing' block combines controller commands from the `Attitude Controller', `Yaw Controller' and `Height Controller' to calculate appropriate rotor speed commands using a linear mixing matrix. 

In the `Forces \& Moments' block we take the rotor speeds and calculate the thrust and torque of each rotor based on the relative airflow through the blades. 
Then the yawing torque will be obtained by simply summing up the torque of each rotor.
Rolling and pitching torques can also be calculated by multiplying the thrust of each rotor with corresponding torque arms.
Meanwhile, we have also introduced a drag term on the fuselage caused by aircraft climb/descent, of which the direction is opposite to the vector sum of aircraft velocity. 
The collective thrust would be equal to the sum of thrust of each rotor combined with the drag effect. 

Afterwards, the thrust and torques are fed to the `Hexacopter Dynamics' block. Assuming the UAV is a rigid body, Newton's second law of motion is used to calculate the linear and angular accelerations and hence the state of the drone will be updated.
To convert the local velocities of the UAV to the earth-based coordinate we will need a rotation matrix, which is parameterized in terms of quaternion to avoid singularities caused by reciprocating trigonometric functions (gimbal lock).

Finally, closed-loop simulations have been tested to validate the functionality of the Simulink model. Tuned PID controllers that display fast response and low steady output error are used in both the inner and outer loops as a challenging benchmark.

\section{Problem Description} \label{sect:prob_desc}

In this work, we are aiming to develop an SNN controller for height control of a hexacopter without explicit modeling of the UAV.
Hebbian plasticity that is evolved offline enables online adaptation to cross the gap between the identified model and the targeted plant.

The controller takes some known states of the plant model (i.e., error in z-axis between the desired and current position as well as the vertical velocity) and learns to generate a functional action selection policy.
The output is a thrust command that will be fed into the plant so that its status can be updated.

Our approach to resolve the problem is threefold. 
First, system identification is carried out to construct a heave model to loosely approximate the dynamics of the hexacopter.
Then neuroevolution is used to search for functional SNN controllers to control the identified heave model.
Network topology and initial weight configurations are determined.
Finally, the fittest controller is selected for further evolution.
Hebbian plasticity is activated so that the network is able to adapt connection weights according to local neural activations.
An EA is used to determine the best plasticity rules by evolving the two parameters $k_m$ and $k_c$ in Eq.~\ref{eq:hebb}.
Each connection can develop its own plasticity rule.
The above-mentioned processes will be offline and only involve the identified model, and the dynamics of the hexacopter are unknown to the controller.

On completion of training, the champion network with the best plasticity rules will be deployed to drive the hexacopter model, which is a more true-to-life representation of the real plant.





\begin{figure}[tb]
    \centering
    \includegraphics[width=0.43\textwidth]{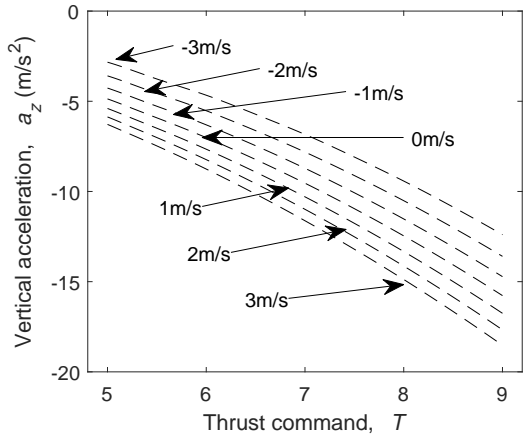}
    \caption{Nonlinear relationship between vertical velocity $v_z$ (-3\,m/s to 3\,m/s), thrust command $T$ and vertical acceleration $a_z$.}
    \label{fig:acc-tvz}
\end{figure}

\section{Identification of Heave Model} \label{sect:sysid}

We first build a loose approximation to resemble the heave dynamics of the hexacopter.
Essentially, this is to model the relationship between the vertical velocity $v_z$, collective thrust $T$ and the vertical acceleration $a_z$.
Fig.~\ref{fig:acc-tvz} shows the nonlinear response of vertical acceleration with varying thrust command when the vertical speed is set as -3\,m/s to 3\,m/s.
Note here that the acceleration is actually the net effect of z-axis force acting on the body, which are generated from the rotor thrust, vertical drag caused by rotor downwash and fuselage.
The net acceleration $a_n$ would be $a_z$ plus the gravitational acceleration $g$. 

\begin{figure}[tb]
    \centering
    \includegraphics[width=0.43\textwidth]{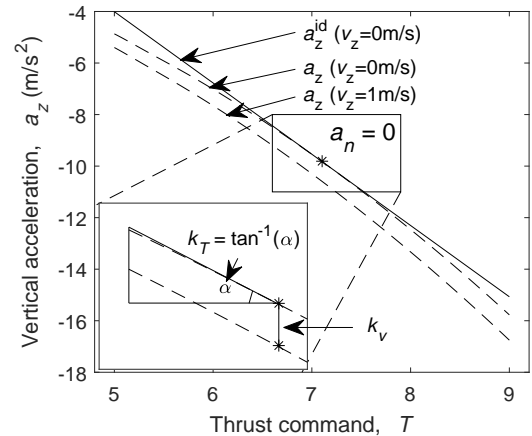}
    \caption{Acceleration curves of the identified model ($a_z^{id}$) and the hexacopter model ($a_z$) with varying thrust command. 
    The identified curve is tangent with that of the hexacopter model at the point where the net acceleration $a_n$ is 0.
    $\alpha$ is the slope angle of the identified linear curve, from which $k_T$ is obtained.
    $k_v$ is calculated from the vertical distance between the two nonlinear curves.
    }
    \label{fig:idmodel}
\end{figure}

In our identified model, vertical acceleration $a_z$ is approximated as a linear combination of the thrust command $T$ and vertical speed $v_z$.
$v_z$, on the other hand, is obtained by integrating the net acceleration of z-axis $a_n$:

\begin{equation}
\begin{aligned}
    a_z &= k_T T + k_v v_z + b \\
    a_{n} &= a_z + g \\
    v_z &= \int a_{n}
    \label{eq:idmodel}
\end{aligned}
\end{equation}
where $k_T$ and $k_v$ are configurable coefficients; $b$ is a bias that is also tunable to make sure that the linear function will be expanded at the point where the net acceleration equals zero, i.e., $a_z = -g$.


We take two of the acceleration curves from Fig.~\ref{fig:acc-tvz} (i.e., for $v_z$ = 0\,m/s and $v_z$ = 1\,m/s) to model the linear function.
The resulting identified linear model is given in Fig.~\ref{fig:idmodel}.
$k_T$ is identified as the slope of $a_z$ against $T$ when $v_z$ = 0 at the point where $a_n = 0$.
$k_v$ is then calculated from the vertical distance between the two nonlinear curves.
Finally, $b$ is set to shift the linear curve vertically, so that the identified model will be tangent with the hexacopter curve at the point where $a_n = 0$.

\begin{figure}[htbp]
    \centering
    \includegraphics[width=0.36\textwidth]{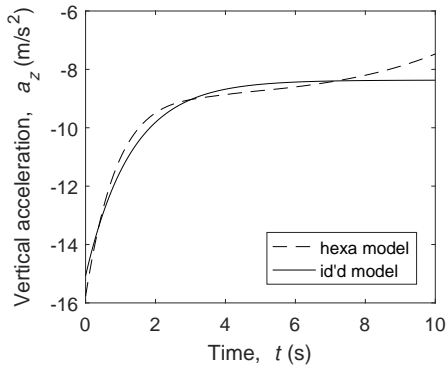}
    \caption{Validation of identified heave model. System response of the two models when fed with the same thrust command signal.}
    \label{fig:valid}
\end{figure}

Finally, the same random thrust command is fed to the two different models for validation of functional similarity. System response of the two models are given in Fig.~\ref{fig:valid}.
Clearly the response of the identified model differs from the hexacopter model, which is desired, but still the identified model approximates the original system.

\section{Controller Development and Deployment} \label{sect:ctrller}

\subsection{Evolution of Non-plastic Controllers}
\label{ssect:non-plastic}

With the identified model we have developed according to Eq.~\ref{eq:idmodel}, we begin to search for optimal network compositions by evolving SNNs using our NEAT implementation.
By `optimal,' we mean the SNN controller is defined to be able to drive the plant model to follow a reference signal with minimal error in height during the course of flight.
Each simulation (flight) lasts 80\,s and is updated every 0.02\,s.

To speed up the evolution process, the whole simulation in this part is implemented in C++, with our \texttt{eSpinn} package.
At the beginning, a population of non-plastic networks are initialized and categorized into different species.
These networks are feed-forward, fully-connected with random connection weights. 
The initial topology is 2-4-1 (input-hidden-output layer neurons), with an additional bias neuron that is connected to all hidden and output layer neurons.
The two inputs consist of error of position in z-axis $e_z$ and vertical velocity $v_z$, other than which, the system's dynamics are unknown to the controller. 
Output of the controller is thrust command that will be fed to the plant model.

Encoding of sensing data is done by the \texttt{encoding} neurons in the input layer. 
Input data are first normalized within the range of [0,1], so that the standardized signal can be linearly converted into a current value (i.e., $I$ in Eq. \ref{eq:izhi}).
This so-called `current coding' method is a common practice to provide a notional scale to the input metrics.

After initialization, each network will be iterated one by one to be evaluated against the plant model. A fitness value will be assigned to each of them based on their performance. 
Afterwards, these networks will be ranked within their species according to their fitness values in descending order. 
A newer generation will be formed from the best parent networks using NEAT:
only the top 20\% of parents in each species are allowed to reproduce, after which, the previous generation is discarded and the newly created children will form the next generation.
During evolution, hidden layer neurons will increase with a probability of 0.005, connections will be added with a probability of 0.01.
Connection weights will be bounded within [-1, 1].

The program terminates when the population's best fitness has been stagnant for 12 generations or if the evolution has reached 50 generations\footnote{empirically determined}.
During the simulation, outputs of the champion will be saved to files for later visualization.
The best fitness will also be saved.
Upon completion of simulation, data structure of the whole population will be archived to a text file, which can be retrieved to be constructed in our later work.

\subsection{Searching Solutions}

Note the control system to be solved is a Constraint Problem \cite{Michalewicz1996}, because the height of the UAV must be bounded within some certain range in the real world. 
However, constraint handling is not straightforward in NEAT  -- invalid solutions that violate the system's boundary can be generated, even if their parents satisfy these constraints. 
Therefore, in this paper we use the feasibility-first principle \cite{Michalewicz1996} to handle the constraints. 

We divide the potential solution space into two disjoint regions, the feasible region and the infeasible, by whether the hexacopter is staying in the bounded area during the entire simulation. 
For infeasible candidates, a penalized fitness function is introduced so that their fitness values are guaranteed to be smaller than those feasible.

We define the fitness function of feasible solutions based on the mean normalized absolute error during the simulation:
\begin{equation}
f = 1- \bar{\lvert e_n \rvert} \label{fit_f}
\end{equation}
where $\lvert e_n \rvert$ denotes the normalized absolute error between actual and reference position. Since the error is normalized, desired solution will have a fitness value close to 1.

For infeasible solutions, we define the fitness based on the time that the hexacopter stays in the bounded region:
\begin{equation}
    f = k (t_i / t_t) \label{fit_if}
\end{equation} 
where $t_i$ is the steps that the hexacopter successively stays in the bounded region, and $t_t$ is the total amount of steps the entire simulation has.
Penalty is applied using a scalar $k$ of 0.2.

\subsection{Enabling Plasticity}

Once the above step is done to discover the optimal network topology, we proceed to consider the plasticity rules.
The champion network from the previous step is loaded from file, with the Hebbian rule activated.
It is spawned into a NEAT population, where each network connection has randomly initialized Hebbian parameters (i.e., $k_m$ and $k_c$ in Eq.~\ref{eq:hebb}).

Networks are evaluated as previously stated. 
The best parents will be selected to reproduce.
During this step, all evolution is disabled except for that of the plasticity rules, e.g.
the EA is only used to determine the optimal configuration of the plasticity rules.


Upon completion of the previous steps, the final network controller is obtained and ready for deployment.
To construct the controller in the Simulink hexacopter model, it is implemented as a C++ S-function block.

\section{Results and Analysis} \label{sect:results}

10 runs of the controller development process have been conducted to perform statistical analysis.
Data are recorded to files and analyzed offline with MATLAB.

\subsection{Adaptation in Progress}
Table \ref{tab:fit} shows the fitness changes of the best controller during the course. 
From left to right are non-plastic networks controlling the identified model, plastic networks controlling the identified model and plastic networks controlling the hexacopter model, respectively.
The fitness values are averaged among the 10 runs.

\begin{table}
    \caption{Best Networks' Mean Fitness Values in Progress}
    \label{tab:fit}
    \begin{tabular}{cccc}
        \toprule
        & \multirow{2}{*}{\parbox{0.24\linewidth}{\centering Non-plastic on id'd model}} &
        \multirow{2}{*}{\parbox{0.24\linewidth}{\centering Plastic on id'd model}} & 
        \multirow{2}{*}{\parbox{0.24\linewidth}{\centering Plastic on hexa model}} \\
        \\ \midrule
        Fitness & 0.9189 & 0.9349 & 0.9298 \\
        \bottomrule
    \end{tabular}
\end{table}

As stated in \ref{ssect:non-plastic}, evolution would be terminated if the performance does not improve for 12 consecutive generations before the threshold of 50. 
For non-plastic controllers, only one of the 10 runs has reached the threshold, and its fitness has only increased by 0.0034 in the last 15 generations.
This indicates the evolutionary runs of non-plastic controllers have plateaued and further evolution is unlikely to find better solutions.
On the other hand, when plasticity is enabled, an increase in fitness can be clearly observed when controlling the same identified model.
The plastic controllers demonstrate better performance even when transferred to control the hexacopter model that has different dynamics.

\subsection{Plastic vs. Non-plastic}

\begin{table}
    \caption{Fitness of Non-Plastic vs. Plastic Controllers on the Hexacopter Model}
    \label{tab:p-vs-np}
    \begin{tabular}{ccc}
        \toprule
        Fitness & Non-plastic & Plastic \\
        \midrule
        Run 1 & 0.9188 & 0.9350 \\
        Run 2 & 0.9074 & 0.9271 \\
        Run 3 & 0.9261 & 0.9396 \\
        Run 4 & 0.9280 & 0.9465 \\
        Run 5 & 0.9053 & 0.9162 \\
        Run 6 & 0.9046 & 0.9166 \\
        Run 7 & 0.9174 & 0.9338 \\
        Run 8 & 0.9188 & 0.9256 \\
        Run 9 & 0.9219 & 0.9366 \\
        Run 10 & 0.9210 & 0.9207 \\
        \midrule
        Mean  & 0.9169 & 0.9298 \\
        \bottomrule
    \end{tabular}
\end{table}

A second comparison is conducted between non-plastic and plastic controllers on the hexacopter model.
Results are given in Table \ref{tab:p-vs-np}.
For 9 out of the 10 runs, we can see a performance improvement when plasticity is enabled.
The only one not being better, still has a close fitness value.
Statistic difference is assessed using the two-tailed Mann-Whitney \emph{U}-test between the two sets of data.
The $U$-value is 21, showing the plastic controllers are significantly better than the non-plastics at $p < 0.05$.

\begin{figure}[tbp]
    \centering
    \includegraphics[width=0.47\textwidth]{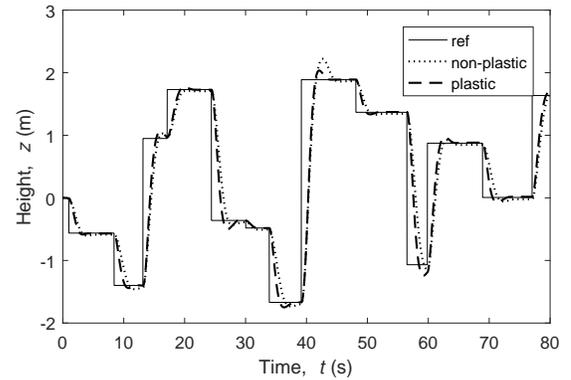}
    \caption{Height control using the non-plastic and plastic SNNs}
    \label{fig:np-vs-p}
\end{figure}

Fig. \ref{fig:np-vs-p} shows a typical run using the non-plastic and plastic controller.
We can see the plastic control system has a faster response as well as smaller steady error.
It is clear that plasticity is a key component to bridge the gap between two models.

\subsection{Validation of Plasticity}

\begin{figure*}[htbp]
    \centering
    \includegraphics[width=0.92\textwidth]{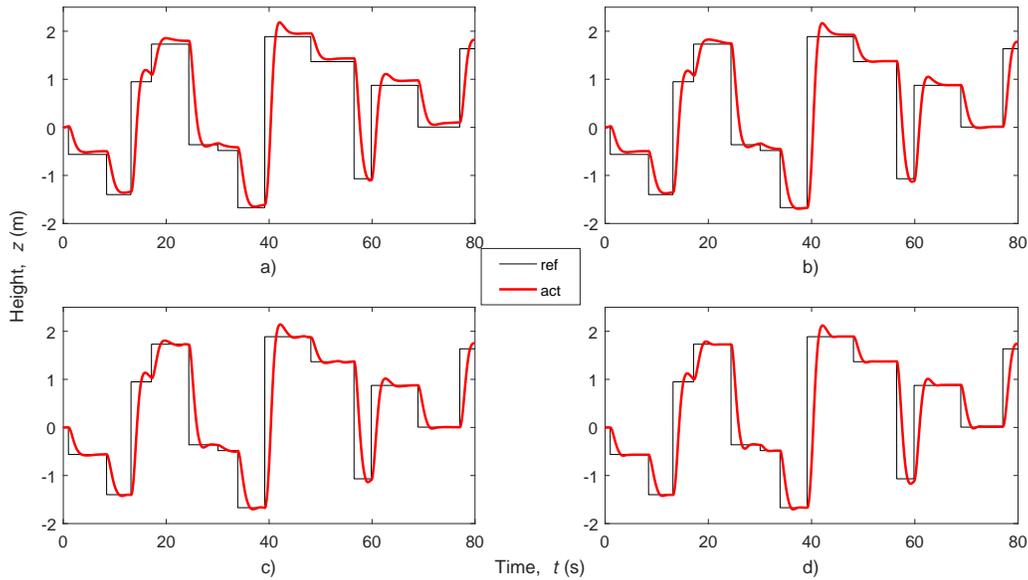}
    \caption{Performance improvement during 4 consecutive runs when a) plasticity is disabled; b-d) plasticity is enabled}
    \label{fig:np_p123}
\end{figure*}

To verify the contribution of the proposed Hebbian plasticity, we extract the evolved best plastic rule and applied it to other networks that have sub-optimal performance.
With plasticity enabled, a sub-optimal network is selected to repetitively drive the hexacopter model to follow the same reference signal.
Fig. \ref{fig:np_p123} shows the progress of 4 consecutive runs when a) plasticity is disabled; b-d) plasticity is enabled.

We can see that in Fig. a), there is a considerable steady system output error.
When plasticity is turned on, connection weights begin to adjust themselves gradually. 
The system follows the reference signal with a decreasing steady error until around 0.005\,m.
Meanwhile a fitness increase is witnessed from a) 0.921296, b) 0.927559, c) 0.932286 to d) 0.933918.

The same results can be obtained when the rule is assigned to other near-optimal networks, while for those with poor initial performance, plasticity learns worse patterns.
This analysis has justified our evolutionary approach to search for the optimal plastic function, demonstrating that plasticity narrows the reality gap for evolved spiking neurocontrollers.


\subsection{Comparing with PID control}

\begin{figure}
    \centering
    \includegraphics[width=0.47\textwidth]{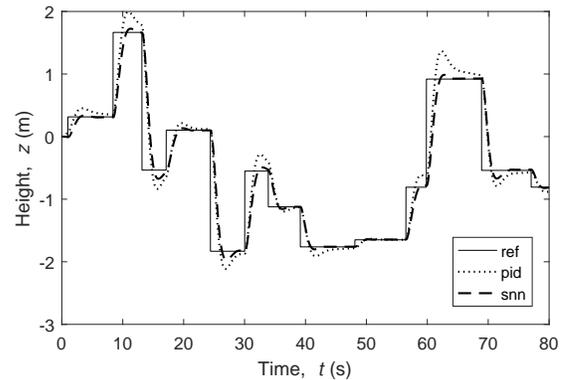}
    \caption{Height control using PID and plastic SNNs}
    \label{fig:pid-vs-p}
\end{figure}

PID control is a classic linear control algorithm that has been dominant in engineering.
The aforementioned PID height controller is taken for comparison.
Note here the PID controller is designed directly based on the hexacopter model, 
whereas the SNN controller only relies on the identified model and utilizes Hebbian plasticity to adapt itself to the new plant model.
System outputs of the two approaches is given in Fig. \ref{fig:pid-vs-p}. Evidently our controller has smaller overshoot and steady output error.
The PID controller has a mean absolute error of 0.108\,m during the course of flight, while our plastic SNN controller has a value of 0.090\,m.

\section{Discussion}
\label{sect:discuss}

When transferring the pseudo-optimal controllers to physical-world applications, one may argue we can still rely on evolution to tweak the connection configurations. 
However, one main problem is that learning in evolution cannot be continuous because the fitness signal during the process is not immediately available.
What we propose here is to evolve in advance the adaptive characteristics of the neurocontroller, such that the controller can be self-organizing and adaptive to model changes during the entire lifetime in real-time.
There is no guarantee that any Hebbian rules can perform synaptic changes on the desired direction. That is why we use evolution to discover functional Hebbian rules in which synapses build up over time in a meaningful manner.


\section{Conclusions} \label{sect:conclusion}

Our work has presented a solution to applied evolutionary aerial robotics,
where evolution is used not only in network initial construction,
but also to formulate plasticity rules which govern synaptic self-modulation for online adaptation based on local neural activities.
We have shown that plasticity can make the controller more adaptive to model changes in a way that evolutionary approaches cannot accommodate.
We are currently in the process of applying this controller development strategy to a real hexacopter platform, and expanding from height control to encompass all degrees of freedom in the UAV.




\end{document}